**Title:** Domain-guided data augmentation for deep learning on medical imaging

**Authors:** Chinmayee Athalye MS[1], Rima Arnaout MD[2]*

[1]Department of Medicine, Division of Cardiology

Bakar Computational Health Sciences Institute

University of California, San Francisco

[2]Department of Medicine, Division of Cardiology

Bakar Computational Health Sciences Institute

Chan Zuckerberg Biohub Intercampus Research Award Investigator

Biological and Medical Informatics Graduate Program

University of California, San Francisco

Email: rima.arnaout@ucsf.edu

* Corresponding Author

# Abstract

While domain-specific data augmentation can be useful in training neural networks for medical imaging tasks, such techniques have not been widely used to date. Here, we test whether domain-specific data augmentation is useful for medical imaging using a well-benchmarked task: view classification on fetal ultrasound FETAL-125 and OB-125 datasets. We found that using a context-preserving cut-paste strategy, we could create valid training data as measured by performance of the resulting trained model on the benchmark test dataset. When used in an online fashion, models trained on this data performed similarly to those trained using traditional data augmentation (FETAL-125 F-score 85.33 ± 0.24 vs 86.89 ± 0.60, p-value 0.0139; OB-125 F-score 74.60 ± 0.11 vs 72.43 ± 0.62, p-value 0.0039). Furthermore, the ability to perform augmentations during training time, as well as the ability to apply chosen augmentations equally across data classes, are important considerations in designing a bespoke data augmentation. Finally, we provide open-source code to facilitate running bespoke data augmentations in an online fashion. Taken together, this work expands the ability to design and apply domain-guided data augmentations for medical imaging tasks.

# Introduction

First demonstrated for non-medical tasks, deep learning has shown remarkable utility for medical imaging in recent years (1–3) – with little to no need to adapt neural network architectures for medical domains. Critical to training a robust and generalizable deep learning model is the use of augmentation techniques to enhance the diversity of available training data (4). Traditional data augmentations for images include shear, rotation, flipping, blurring, contrast stretching, and other operations (5) performed in an online manner during training.

While neural network architectures can be applied off-the-shelf to medical imaging tasks, the same is not always true for data augmentation techniques for two reasons. First, domain expertise is needed to apply even traditional data augmentations correctly. For example, flipping X-rays can change the laterality of the image; rotating images too much can change the training labels in adult echocardiography (6) but can be very helpful in fetal ultrasound (1); and too much distortion can affect image quality in ways that are not clinically relevant or can obscure important anatomic structures. The spectrum of traditional augmentations/hyperparameters that can be used in a given medical imaging task is therefore constrained, and model training can face data starvation.

Second, traditional data augmentations do not fully exploit the domain-specific traits of medical images. At best, this means that opportunities to further expand the training dataset are not leveraged. At worst, failing to apply domain-specific data augmentation can lead to fundamental bias in trained models, such as relying on fiducial markings associated with skin biopsies predict skin cancer (7), or relying on presence of endotracheal tubes or other equipment to predict severity of disease from chest X-rays (8). To fully exploit domain-specific traits, domain-guided data augmentation can be useful for medical imaging, but it has not been widely used to date due to complexity of implementation compared to traditional methods.

In this paper, we present a custom, context-preserving, anatomy-aware way of combining two images in an online fashion to create new training data for medical imaging. To illustrate the development and evaluation of domain-guided data augmentation, we used a well-benchmarked task of fetal cardiac view classification (1), where five screening views of the fetal heart—called 3-vessel trachea (3VT), 3-vessel view (3VV), left-ventricular outflow tract (LVOT), axial 4-chamber (A4C), or Abdomen (ABDO)—must be distinguished from non-target (NT) images. Not only is there imbalance among view classes (see Methods), but each frame required painstaking labeling, both motivating a desire to augment the training dataset.

The defining features of each screening heart view are found within the axial thorax, while features outside the thorax (e.g. arms, legs, umbilical cord) are non-specific. We therefore chose cutting-and-pasting thoraces from one image into another as our domain-guided data augmentation strategy. Cut-paste for a deep learning application was first used for instance detection (9) and has been shown to improve performance (9–12); some suggest that random positioning (9,12) of pasted objects performs well, while others have shown that context-aware approaches (10) are important. CutMix (13) and Mixup (14) are cut-paste approaches that have been used in non-medical settings, but neither creates visually realistic images. For medical imaging, cut-paste data augmentation has been used in an offline manner to generate instances with lesions in chest CT scans (15). However, this method used a complex blending tool based on Poisson image editing which cannot be scripted to generate images in an online manner, where online data augmentation provides regularization which is critical to training neural networks. TumorCP used an online cut-paste strategy for the task of kidney tumor segmentation (16). However, the test set was small (only 42 images) and performance variable (up to 28% standard deviation in reported Dice score). In our case, cutting the thorax from one image and pasting it into another preserves anatomical context, and new images do not require relabeling by clinicians.

With this use case, we tested whether domain-specific data augmentation was useful for deep learning tasks in medical imaging and report on design and performance, compared to traditional data augmentation.

## Methods

*Dataset:* The training data consists of still frames from fetal screening ultrasounds and fetal echocardiograms, as in (1). This dataset is imbalanced across the six classes – 3193 (6.1% of total training data) 3VT, 6178 (11.8%) 3VV, 6735 (12.9%) LVOT, 6029 (11.5%) A4C, 5206 (9.9%) Abdo, and 25082 (47.8%) NT images.

Two test datasets were used. The FETAL-125 test set (1) consists of 306 (2.7% of this test set) 3VT, 890 (7.8%) 3VV, 1800 (15.7%) LVOT, 3521 (30.8%) A4C, 563 (4.9%) ABDO, and 4365 (38.1%) NT images for a total of 11,445 images. The original OB-125 test set (1) consists of 678

(0.3%) 3VT, 2431 (1.1%) 3VV, 3755 (1.7%) LVOT, 16852 (7.6%) A4C, 3473 (1.6%) ABDO, and 193801 (87.7%) NT images. This vast class imbalance in OB-125 makes it hard to use any changes in the F-score or accuracy for comparing performance between experiments. Hence, we randomly sampled a subset of the OB-125 that had 678 images for each target view class and 3390 NT images; this 'balanced OB-125' test set was used.

*Hybrid image generation*: The workflow to generate a cut-pasted hybrid image is described in **Figure 1a**. We used a thorax detector segmentation model described in (1) to extract the region of interest in each image. We used the convex hull of the segmented thorax region, closing any holes in the segmented region and removing any extraneous pixels. To make the extraction and combination appear seamless, we approximated the thorax to its best fit circular region (an axial thorax should be round), rejecting highly eccentric (> 0.75 eccentricity) segmentations. We define the eccentricity as the ratio of the major axis to minor axis of the convex hull. We then used a binary mask to separate the image into two components – the thorax (donor) and the background image with a cavity (acceptor) in place of the thorax as shown in **Figure 1a**. All images from which we can extract a valid donor and acceptor are deemed cut-paste eligible. Hybrid images were then created by randomly choosing a donor and an acceptor, resizing the donor thorax to the size of the acceptor cavity, randomly rotating the thorax 10-350 degrees, and pasting the donor region in the acceptor cavity. The new hybrid image carried the same class label as the donor thorax. All these steps were performed using Python 3.6, with scikit-image version 0.16.2 and OpenCV version 4.5.1 packages.

*Training parameters*: For all experiments, the following architecture and training parameters were used. We used the ResNet architecture (17) for the view classifier with an input image size of 80 x 80 pixels. The batch size was set to 32. We used the Adam optimizer with a starting learning rate of 0.0005 and then adaptively reduced the learning rate by a factor of 0.5 if there was no improvement > 0.001 in the validation loss for 5 consecutive epochs. The maximum number of epochs was set to 250 with early stopping if there was no improvement > 0.001 in the validation loss for 15 consecutive epochs. Dropout of 50% was applied before the final fully connected layer. For statistical rigor, three replicates were performed for each experiment (18–20). All models were implemented in Python 3.6 and TensorFlow 2.1.3 Keras 2.3.0 framework. We executed the training on an AWS EC2 'g4dn.4xlarge' instance with a Nvidia T4 GPU and 64GB memory.

*Traditional data augmentation*: For all experiments where traditional data augmentation was used, online data augmentation consisted of applying Gaussian blur and rescaling the image intensity to within $2^{nd}$ and $98^{th}$ percentile with 50% probability each, rotating images for up to 10º, width and height shift of up to 20% of original size, zooms of up to 50%, and vertical and horizontal flips at random.

*Statistical testing:* We used two-tailed t-tests to compare performance between experiments except where specified.

# Results

To test whether a domain-guided data augmentation approach could produce valid training data for deep learning applications, we performed several experiments using the thorax cut-paste augmentation strategy on fetal ultrasound images as a use case, evaluating using the benchmark task of 6-view classification with ResNet. We evaluated the models on two test sets: (1) fetal echocardiograms from 125 patients formed the FETAL-125 test set and (2) a view-balanced subset (see Methods) of the corresponding patients' obstetric ultrasound screenings formed the OB-125 balanced test set.

*Cut-paste technique produces images that are valid and are useful in training*

We first tested whether our hybrid image generation method could produce valid training data by synthesizing images offline, and then using them to re-train our benchmark classification task. Using our pipeline (Methods, **Figure 1a**), we created new hybrid images that were realistic overall with only small cut-paste combination artifacts **(Figure 1b)**. Adjacent frames in an ultrasound video from the same patient will generally have very similar heart structure (21), so we randomly chose only one frame per view per patient ID for target views and five frames per patient ID for NT view to generate a set of candidate images; these frames also had to pass quality control (**Methods**). Thus, we used *only five percent* of the original training data to synthesize thousands of new images such that the new hybrid training dataset was approximately the size of the original training dataset **(Table 1)**.

To prevent the model from overfitting on any combination artifacts, all view classes had an approximately equal percentage of hybrid images: about 90%. This meant that the classes with lower numbers of donors were sampled more times than others **(Table 1)**. The remaining 10 percent of images were original (non-hybrid). Overall, the entire hybrid training dataset was derived from only 13.7 percent of the original images (**Table 1**).

We re-trained our model using the hybrid training dataset described above and the hyperparameters and traditional data augmentation as in the Methods. The model trained on the new data set without overfitting. **Table 2** shows that accuracy and F-score for the models trained on this hybrid dataset is comparable with the original model (FETAL-125 F-score and accuracy p-values both 0.45; OB-125 F-score and accuracy p-values both 0.12). As with the original dataset

(1), this hybrid-trained model performs slightly better on FETAL-125 (higher-quality fetal echocardiogram images) compared to OB-125 (screening ultrasound images) (F-score 91.99 ± 2.62 vs 80.76 ± 0.99, accuracy 97.33 ± 0.87 vs 93.58 ± 0.33; both p-values 0.002). By recapitulating original model performance using a dataset overwhelmingly composed of hybrid images, we demonstrate that the cut-paste strategy is a viable method to generate additional training images.

*When used in an online manner, the cut-paste technique provides the regularization needed to train a neural network*

Online data augmentation acts as a regularization technique while training a neural network (4). In the previous experiment, the hybrid images were shown to serve as valid training data, while regularization was provided from online traditional data augmentation. We next tested whether, when performed in an online manner, hybrid images could also provide the necessary regularization to avoid overfitting (without need for traditional data augmentation). We used all cut-paste eligible images from the training data, or 80.1% of the original training images **(Table 3)**. Implementing the cut-paste strategy in an online manner generates a new set of hybrid images per epoch due to the random combinations. With our cut-paste eligible data (**Table 3,** total number of eligible donors), it is possible to generate 1.2e9 new unique hybrid images.

**Figure 2a** compares the training loss progress of the model when the cut-paste strategy is used as an online data augmentation compared to traditional (online) data augmentation. The training loss when no data augmentation is used is shown as a negative control, showing that the model quickly overfits the training data when no data augmentation is used. Comparison of the training loss plots show that the cut-paste strategy works better (p-value 4.92 10e-5) as an online data augmentation strategy. **Figure 2b** shows the F-score and accuracy values of this newly trained model for comparison.

*The ability to apply bespoke data augmentation equally across classes is an important design consideration*

While the above experiment showed better training loss using the cut-paste data augmentation strategy, the confusion matrices in **Figures 2c-2d** give a more nuanced view of the test performance, demonstrating a skew toward prediction of the NT class. We hypothesized this was due to an imbalance in how the chosen bespoke data augmentation can be applied to images of different classes; only 41 percent of NT images were cut-paste eligible compared to 85-93 percent for the other view classes **(Table 3)**, because the NT class includes many images where no thorax appears.

When we implemented cut-paste strategy as an online data augmentation with probability 1 (i.e., every image eligible to undergo cut-paste processing would receive it), it meant that each batch of training data would contain only up to 15 percent of target-class images in their original form with the rest as hybrid images. But due to the imbalance in eligible images in the NT class, the model would see 59% of the NT-class images in their original form during training. The test images are all original unchanged images, and the model seems to overfit on this characteristic. Hence, the results in Figure 2 cannot be used to make a direct comparison of cut-paste data augmentation to our original model or traditional data augmentation.

We mitigate this imbalance of hybrid and unchanged training data using a sampling strategy in the subsequent experiments to include similar proportions of online data-augmented and non-data-augmented images in each training batch, regardless of whether traditional or bespoke data augmentation is used, for a more fair comparison of the two methods.

*Cut-paste data augmentation performs similarly to traditional data augmentation in a classification task*

To counter the class imbalance in application of the cut-paste technique while preserving online data augmentation, we employed a new sampling strategy for image batches during online training. All the target class cut-paste eligible images (85-93%) are passed as both hybrid images and in their original unchanged form to the model at every epoch. For the NT class, the cut-paste eligible images (41%) are passed only in the hybrid form and the rest (59%) in their original, unchanged form.

Similarly, for training using traditional data augmentation, images were sampled such that about 50% of images per training batch underwent traditional data augmentation, while the rest of the images in the batch remained unchanged. Therefore, for each view class, target classes and NT class alike, each training batch contained approximately 50% augmented and 50% original images, regardless of whether traditional or bespoke data augmentation was applied. The sampling strategy is demonstrated for a toy batch of data in **Supplementary table 1**.

The overall and per-class test performance for these experiments are given in **Table 4** and **Table 5**. The new sampling strategy shows improvement in performance over the previous experiment (**Figure 2b**) as is seen in the improved F-score and accuracy over both test sets in **Table 4** (all p-values < 0.01). On the FETAL-125 test set, absolute performance from traditional and bespoke data augmentation methods is quite similar (e.g. accuracy of 95.63 ± 0.20 with traditional data augmentation vs. 95.11 ± 0.08 with bespoke data augmentation), despite statistical significance weighing in favor of traditional augmentation. On the OB-125 test set, both absolute performance and statistical significance favor bespoke data augmentation (F-score p-value 0.0039, accuracy p-value 0.0016).

The training loss plots for both these experiments and their replicates are shown in **Supplementary figure 1**, showing that the model trained with traditional data augmentation overfit the training data which consists of majority fetal echocardiogram images. This could explain the model's better performance on the FETAL-125 test set, which consists of all fetal echo images, compared to OB-125. Additionally, the changes in per-class recall values for both these experiments are reported in Table 5. We note that the model trained with cut-paste data augmentation outperforms the model trained with traditional data augmentation on 3VT (FETAL-125 recall p-value 0.2189; OB-125 recall p-value 0.0573) which performed the worst with our original model.

# Discussion

In this work, we explore the utility of using domain knowledge in medical images to design bespoke data augmentations for neural network training. The fetal view classification use case provides a helpful demonstration of strategies and potential pitfalls in design, implementation, and benchmarking, due to the size and composition of the available training datasets and the statistical rigor provided in the experiments presented.

Especially when the ability to apply traditional data augmentation may be limited, bespoke data augmentation has the potential to generate thousands of new labeled training images from a limited amount of labeled data. As shown above, training using our hybrid image data produced comparable test performance with only 13.7 percent of data used in the original model. Training data efficiency, in turn, can pay dividends by lightening the data labeling burden, especially when combined with other strategies for training dataset curation (21). This is particularly advantageous in the medical domain where there is a scarcity of data and experts to label this data.

Furthermore, bespoke data augmentation is a valid strategy to mitigate class imbalance in available training data. In all experiments, the model trained on hybrid data showed improved performance for the 3VT class which had the least amount of training data.

Bespoke data augmentation can be implemented online during model training or offline before training. In an online manner, our chosen bespoke augmentation, cut-paste, provides both data augmentation and training regularization, avoiding overfitting even when no other data augmentation is used. Offline implementation provides more granular control over data sampling strategy and seeds can be used for the random combinations to ensure reproducibility. We also make available code for online bespoke data augmentation that can be adapted to the user's augmentation of choice.

With respect to the particular bespoke augmentation chosen in this paper, design of the cut-paste strategy comes with a caveat as it is heavily dependent on the effectiveness of the thorax detector. The performance of the thorax detector is not uniform across all the classes. For NT class especially, there are many images which have no thorax in the frame. The quality control rules do a good job of keeping out the outliers from the thorax detector output. However, these are not foolproof. The imbalance of the original training data combined with the class specific performance of the thorax detector make this strategy not optimal to use for the six-view classifier. However, the class imbalance also reflects the real-world class distribution of fetal ultrasound and echo images and we have tried to accommodate it as well as possible while maintaining experimental rigor to prove our hypotheses. For future work, it would be best to consider domain-guided strategies that can be applied uniformly across all classes for optimal results.

Overall, we find that the ability to design and implement bespoke data augmentations for deep learning tasks in medical imaging expands the researcher's toolbox for training models that are high-performance, clinically relevant, and data-efficient.

# Code Availability

Code will be available at https://github.com/ArnaoutLabUCSF/cardioML

# Author Contributions

R.A. and C.A. conceived of the experiments. C.A. performed experiments and analyses and wrote the manuscript, with critical input from R.A.

# Disclosures

None.

# Funding

R.A. and C.A. are supported by the National Institues of Health (R01HL150394) and the Department of Defense (PR181763). R.A. is additionally supported by the Gordon and Betty Moore Foundation and is a Chan Zuckerberg Intercampus Research Awardee.

# Tables

## Table 1 - Distribution of training data with 90% hybrid images.

| View | 1. Number of original training images | 2. Number of donors | 3. Number of times each donor is sampled | 4. Number of hybrid images generated | 5. Number of original training images sampled | 6. Total number of training images | 7. Percent of hybrid images in total training data |
|---|---|---|---|---|---|---|---|
| 3VT | 3193 | 297 | 22 | 6534 | 700 | 7234 | 90.3 |
| 3VV | 6178 | 439 | 15 | 6585 | 700 | 7285 | 90.4 |
| A4C | 6029 | 570 | 12 | 6840 | 700 | 7540 | 90.7 |
| LVOT | 6735 | 469 | 14 | 6566 | 700 | 7266 | 90.4 |
| Abdo | 5206 | 475 | 14 | 6650 | 700 | 7350 | 90.5 |
| NT | 25082 | 633 | 50 | 31650 | 3500 | 35150 | 90.0 |
| TOTAL | 52423 | 2883 | N/A | 64825 | 7000 | 71825 | 90.2 |

Distribution of training data where 90 percent of training data consists of hybrid images generated in an offline manner (columns 6, 7). Only about five percent (column 2) of the original training images (column 1) are used to generate all the new hybrid training images (column 4). Overall, the entire hybrid training dataset (column 6) was derived from only 13.7 percent of the original images. Column 3 denotes the number of times each donor is samples. A total number is not applicable here.

## Table 2 - Performance of model trained on 90% hybrid data.

| Metric | Fetal-125 | | | OB-125 balanced subset | | |
|---|---|---|---|---|---|---|
| | Original model | 90% hybrid training data | p-value | Original model | 90% hybrid training data | p-value |
| Accuracy | **97.80** | 97.33 ± 0.87 | 0.45 | 93.09 | **93.58 ± 0.33** | 0.12 |
| F-score | **93.40** | 91.99 ± 2.62 | 0.45 | 79.27 | **80.76 ± 0.99** | 0.12 |

Original model was benchmarked in (1). p-values reported using the two-tailed t-test.

Table 3 - Number and class distribution of cut-paste eligible images.

| View | Number of original training images | Number of eligible donors and acceptors | % of original data with eligible donors |
|---|---|---|---|
| 3VT | 3192 | 2721 | 85.24 |
| 3VV | 6178 | 5482 | 88.73 |
| A4C | 6029 | 5612 | 93.08 |
| LVOT | 6735 | 5770 | 85.67 |
| Abdo | 5970 | 5206 | 87.20 |
| NT | 25082 | 10239 | 40.82 |
| TOTAL | 52423 | 35030 | 66.82 |

All hybrid eligible images contribute both a donor thorax and an acceptor cavity.

Table 4 - Comparison of traditional and cut-paste data augmentation approaches when application of augmentation during training time is balanced.

| Metric | Fetal-125 | | | OB-125 balanced subset | | |
|---|---|---|---|---|---|---|
| | Model trained with traditional data augmentation | Model trained with cut-paste data augmentation | p-value | Model trained with traditional data augmentation | Model trained with cut-paste data augmentation | p-value |
| Accuracy | **95.63 ± 0.20** | 95.11 ± 0.08 | 0.0139 | 90.59 ± 0.21 | **91.53 ± 0.04** | 0.0016 |
| F-score | **86.89 ± 0.60** | 85.33 ± 0.24 | 0.0139 | 72.43 ± 0.62 | **74.60 ± 0.11** | 0.0039 |

p-value reported using the two-tailed t-test.

Table 5 – Per-class recall values for results in Table 4

| View | Fetal-125 | | OB-125 balanced subset | |
|---|---|---|---|---|
| | Model trained with traditional data augmentation | Model trained with cut-paste data augmentation | Model trained with traditional data augmentation | Model trained with cut-paste data augmentation |
| 3VT | 62.28 ± 1.75 | **64.49 ± 1.96** | 45.64 ± 2.09 | **49.24 ± 1.09** |
| 3VV | 72.10 ± 2.24 | **74.76 ± 1.40** | 68.90 ± 2.47 | **80.04 ± 2.73** |
| A4C | **86.27 ± 2.50** | 82.67 ± 1.33 | 79.16 ± 1.24 | **80.92 ± 1.70** |
| LVOT | **77.28 ± 1.28** | 76.52 ± 3.84 | **65.68 ± 0.91** | 62.15 ± 0.35 |
| Abdo | 84.08 ± 1.84 | **85.62 ± 2.58** | **79.65 ± 1.67** | 79.15 ± 3.63 |
| NT | **95.95 ± 0.30** | 94.68 ± 0.35 | 77.03 ± 1.21 | **78.96 ± 1.24** |

# Figures

Figure 1 – Workflow for applying a bespoke, cut-paste data augmentation to images.

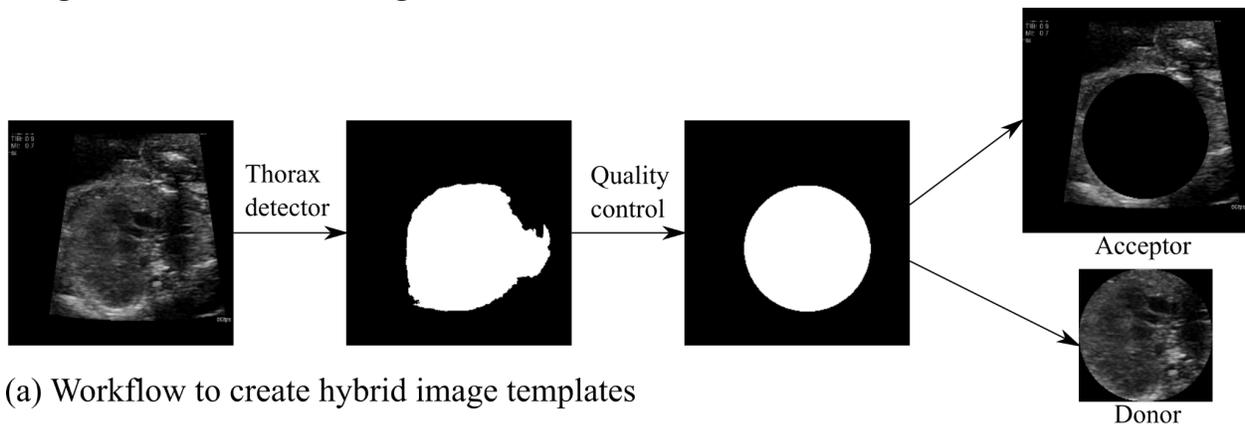

(a) Workflow to create hybrid image templates

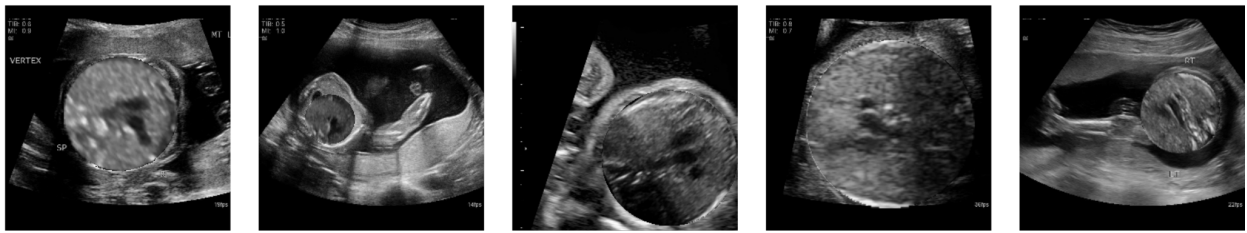

(b) Examples of hybrid images

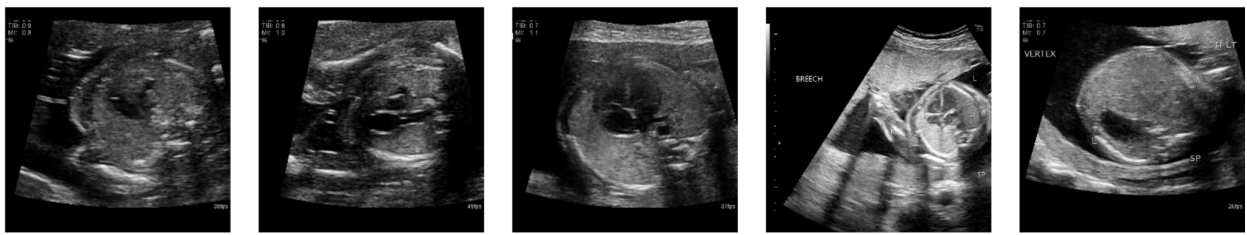

(c) Examples of original images

Figure 1 - (a) Workflow to generate the donor and acceptor templates from a training image. (b) Examples of hybrid images created by random combinations of donors and acceptors. (c) Examples of original, unchanged training images for comparison to hybrid images.

# Figure 2 - Cut-paste as a data augmentation strategy.

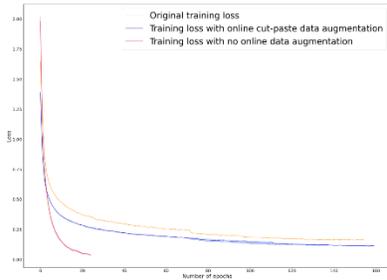

(a) Training loss plots

(b) Performance of the original model and model trained with cut-paste as online data augmentation

| Metric | Fetal-125 | | | OB-125 balanced subset | | |
|---|---|---|---|---|---|---|
| | Original | Online cut-paste | p-value | Original | Online cut-paste | p-value |
| Accuracy | **97.80** | 94.37 ± 0.28 | 0.002 | **93.09** | 91.47 ± 0.27 | 0.009 |
| F-score | **93.40** | 83.11 ± 0.83 | 0.002 | **79.27** | 74.41 ± 0.82 | 0.009 |

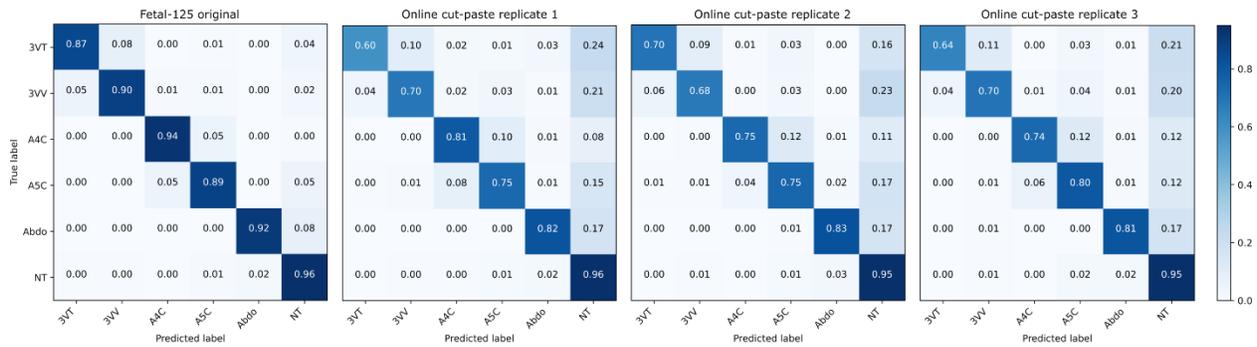

(c) Normalized confusion matrices for Fetal-125 test set

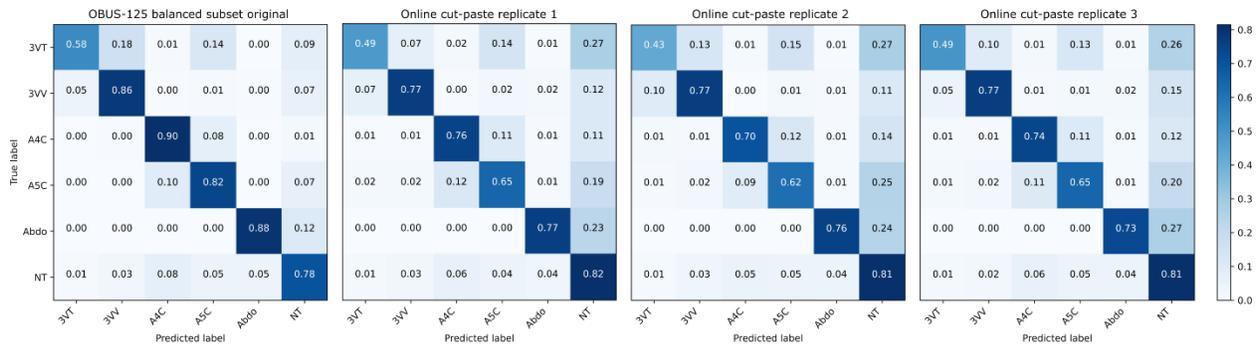

(d) Normalized confusion matrices for OBUS-125 balanced subset

Figure 2 - (a) Loss plots for original training (yellow), three replicates each of - training with no online data augmentation (red, with error as light red), and training with online cut-paste data augmentation (blue, with error as light blue). (c)(d) Normalized confusion matrices for original model and model trained with online cut-paste data augmentation.

# Supplementary materials

## Supplementary table 1 – Example training batch to demonstrate sampling strategy for balancing original and augmented images at training time.

| View | Cut-paste eligible | Cut-paste data augmentation | Traditional data augmentation |
|---|---|---|---|
| 3VT | Yes | Cut-pasted hybrid & unchanged | Traditional DA & unchanged |
| 3VT | No | Unchanged only | Unchanged only |
| 3VV | Yes | Cut-pasted hybrid & unchanged | Traditional DA & unchanged |
| 3VV | Yes | Cut-pasted hybrid & unchanged | Traditional DA & unchanged |
| A4C | Yes | Cut-pasted hybrid & unchanged | Traditional DA & unchanged |
| A4C | Yes | Cut-pasted hybrid & unchanged | Traditional DA & unchanged |
| LVOT | Yes | Cut-pasted hybrid & unchanged | Traditional DA & unchanged |
| LVOT | Yes | Cut-pasted hybrid & unchanged | Traditional DA & unchanged |
| Abdo | No | Unchanged only | Unchanged only |
| Abdo | Yes | Cut-pasted hybrid & unchanged | Traditional DA & unchanged |
| NT | Yes | Cut-pasted hybrid only | Traditional DA only |
| NT | No | Unchanged only | Unchanged only |
| NT | Yes | Cut-pasted hybrid only | Traditional DA only |
| NT | No | Unchanged only | Unchanged only |

If a target view image is cut-paste eligible it is passed in both its hybrid and unchanged form. However, a cut-paste eligible NT image is only passed in its hybrid or Traditional DA form. *DA = Data Augmentation*

# Supplementary figure 1 - Loss plots for training with cut-paste and traditional data augmentation with balanced data

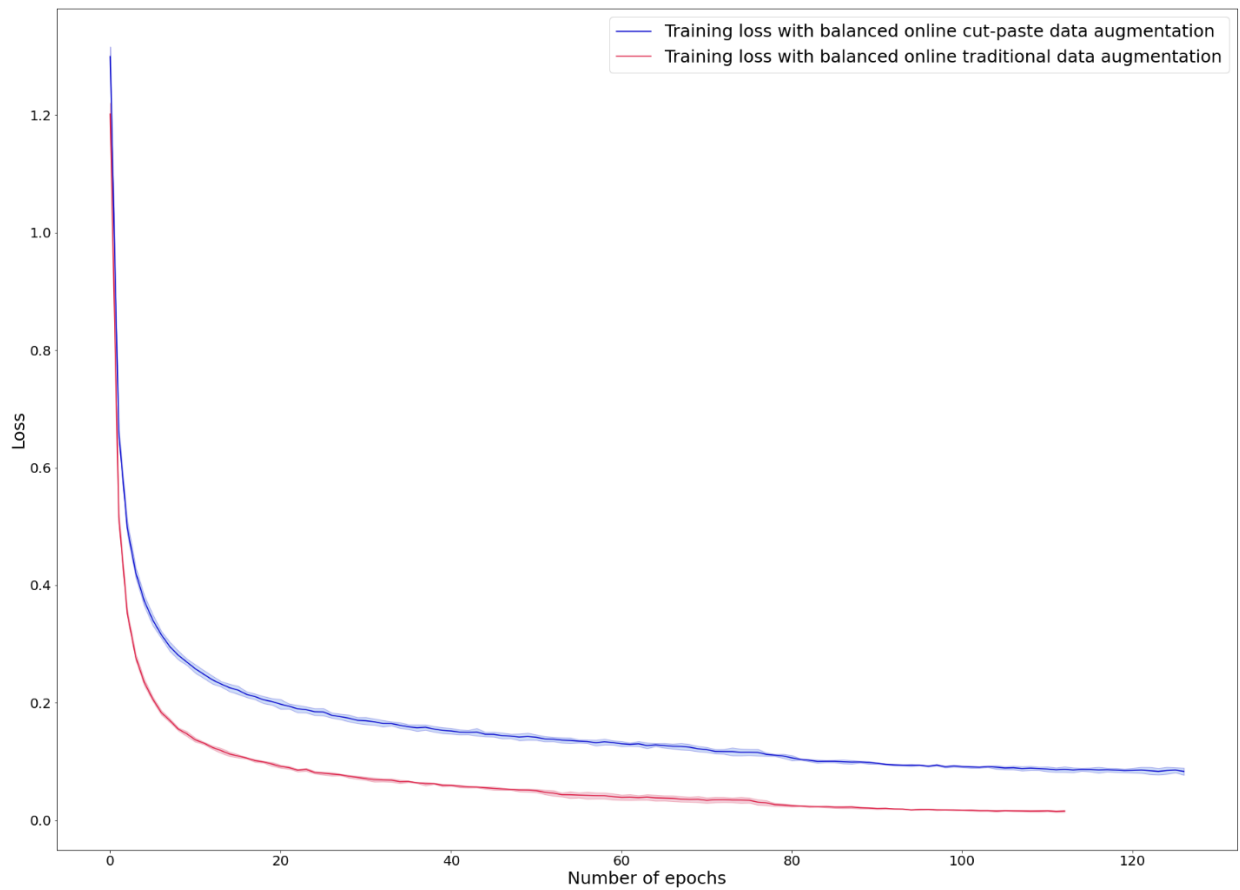

Supplementary figure 1 - Loss plots for training with balanced cut-paste (blue) and traditional (red) data augmentation. Training with traditional data augmentation overfits the training data. We performed three replicates for both (error in light blue and light red, respectively).